# Meta-Unsupervised-Learning:
# A supervised approach to unsupervised learning

Vikas K. Garg and Adam Tauman Kalai

January 3, 2017

**Abstract**

We introduce a new paradigm to investigate unsupervised learning, reducing unsupervised learning to supervised learning. Specifically, we mitigate the subjectivity in unsupervised decision-making by leveraging knowledge acquired from prior, possibly heterogeneous, supervised learning tasks. We demonstrate the versatility of our framework via comprehensive expositions and detailed experiments on several unsupervised problems such as (a) clustering, (b) outlier detection, and (c) similarity prediction under a common umbrella of meta-unsupervised-learning. We also provide rigorous PAC-agnostic bounds to establish the theoretical foundations of our framework, and show that our framing of meta-clustering circumvents Kleinberg's impossibility theorem for clustering.

## 1 Introduction

Unsupervised Learning (UL) is an elusive branch of machine learning, including problems such as clustering and manifold learning, that seeks to identify structure among unlabeled data. Unsupervised learning is notoriously difficult to evaluate, and debates rage about which objective function to use, since the absence of data labels makes it difficult to define objective quality measures. This paper proposes a meta-solution that, by considering the (meta)distribution over unsupervised problems, reduces UL to Supervised Learning (SL). This is a data-driven approach to quantitatively evaluate and design new UL algorithms. Going beyond typical transfer learning, we show how this approach can be used for to improve UL performance for problems of different sizes and from different domains.

As a thought-provoking example, consider clustering the course reviews in Figure 1. The clustering based on sentiment, done by a human, is clearly more "human" than the one based on word count, done by machine. How do humans learn to cluster in this way, and how can computers learn to cluster in this way? People would identify the text as English and may recall related text challenges they have faced, as opposed to image tasks. They may also draw on knowledge about courses. We argue that computers should identify and leverage data from related prior tasks, such as text corpora, clustered documents, clustered reviews, or even clustered course reviews if available. A die-hard K-means advocate may say that the K-means objective of average distance to cluster centers is right, and that the bag-of-words representation used for computing distances is inappropriate. But who should provide the right representation? As computers become increasingly intelligent and human-like, they are being tasked to learn the appropriate representations themselves.

| Sentiment based clustering | Word based clustering |
|---|---|
| The class was great! | **The class was** great! |
| Best course I have taken in a long time. | **The class was** boring. |
|  | **The class was** terrible! |
| I slept the whole time and snored loudly. |  |
| The class was boring. | Best course **I** have taken in a long **time**. |
|  | Waste of **time**. **I** now hate psychology. |
| The class was terrible! | **I** slept the whole **time** and snored loudly. |
| Waste of time. I now hate psychology. |  |

Figure 1: Two different course review clusterings. Left: a sentiment-based clustering, drawing on prior knowledge to establish context. Right: a clustering based on common words and number of words.

Now, the algorithms and experiments in this paper demonstrate that this approach has potential but will not necessarily give good clusterings on this data if problems like this are not well represented in our data. However, our problem framing does allow algorithms to use prior data to perform better UL, just as humans cluster based on knowledge from their own prior experience. This model is intended to capture, to some extent, how humans perform UL and also recent trends in machine learning work on word embeddings and DNN for image classification.

As a motivating example, consider clustering your course reviews. If you were using a popular ML tool, such as scikit-learn [11], you would have to choose among a number of different clustering algorithms, and then you would have to choose how many clusters to have. Is there any principled, data-driven approach to making these decisions? In this paper, we offer a solution using an annotated collection of prior datasets. We present a theoretical model accompanied by empirical results from a repository of classification problems.[1] Each classification problem, after removing labels, is viewed as a clustering problem over the unlabeled data where the (hidden) ground-truth clusters are the sets of examples with each label.

Our model requires a *loss* function measuring the quality of a solution (e.g., a clustering) with respect to some conceptual[2] *ground truth*. Further, we suppose that we have a repository of datasets, annotated with ground truth labels, that is drawn from a *meta-distribution* $\mu$ over problems, and that the given data $X$ was drawn from this same distribution (though without labels). From this collection, one could, at a minimum, learn which clustering algorithm works best, or, even better, which type of algorithm works best for which type of data. The same *meta*-approach could help in selecting how many clusters to have, which outliers to remove, and so forth. With some luck, there might even be a large dataset of course reviews annotated with ratings in $\{1, 2, 3, 4\}$, where a pre-trained classifier could be leveraged.

---

[1] Ideally, we would use a clustering repository such as the *IFCS Cluster Benchmark Data Repository* "datasets with and without given 'true' clusterings," at `http://ifcs.boku.ac.at/repository` once it is sufficiently large.

[2] The ground truth labels may be impossible to acquire for the data at hand.

Our theoretical model, closely related to Agnostic [8] and PAC Learning [19], treats each entire labeled problem analogous to an training example in a supervised learning task. We show how one can provably learn to perform UL as well as the best algorithm in certain classes of algorithms.

Empirically, we run meta-algorithms on the repository of classification problems from `openml.org` which has a variety of datasets from various domains including NLP, computer vision, and bioinformatics, among others. Ignoring labels, each dataset can be viewed as an unsupervised learning problem. Interestingly, we find that this seemingly unrelated collection of problems can be leveraged to improve average performance across datasets. In particular, we first find that the K-means clustering algorithm generally outperforms four others across the collection of problems. While this is not surprising in the sense that K-means is one of the most popular clustering algorithms, our finding is based upon data and not simply "word of mouth." Second, we show that there are systematic *but correctable* biases among a standard heuristic for selecting the number of clusters. Third, we show how the data can be used to decide how many outliers to remove to improve clustering performance. Finally, we also show how to train a neural network using data from multiple classification problems of very different natures to improve performance on a new UL problem.

In some sense, this meta-approach is arguably being used by today's NLP and computer vision practitioners, as well as by humans. For instance, when a human clusters course reviews, they heavily use their knowledge of English that has been already acquired from a number of other learning tasks. Similarly, modern NLP algorithms rely upon Word Embeddings [14] that encode the meanings of words in a constant number of dimensions and are trained on other text corpora or billions of words. In some sense, expecting an algorithm to cluster a small set of text reviews without external data is like asking a person to cluster course reviews written in a foreign language without a dictionary – not only would one have to learn the new language from the reviews themselves but there might not even be enough reviews to adequately cover the language. Similarly, a recent theme in computer vision, given a set of thousands of images, is to re-use pieces of neural networks trained on a labeled set of millions of images.

## 2 Related work

This work relates to and unifies a number of areas within machine learning. In unsupervised learning, the frustrating sense of futility in debating the "best" clustering algorithm is conveyed by Kleinberg's impossibility theorem for clustering [9], though his axioms have been the subject of further debate [20, 1]. In Section 5, we show how meta-clustering circumvents this impossibility in some sense.

Some supervised approaches [4], [6] have been proposed to learn a clustering function, using a dataset where each instance is comprised a set of homogeneous items (such as news articles) and their partitioning into clusters. The learned function could then be used to cluster a new set of items of the same kind. Our approach generalizes the notion by learning across heterogeneous datasets that are compiled from several domains, to find a good clustering for new data.

These domains may have data represented with different dimensionalities.

Our framework can also be viewed as taking an extreme viewpoint of transfer learning. The body of work on (supervised) transfer learning is too large to survey here, see e.g., [10] and references therein. Our work can be viewed as more extreme than typical transfer learning in that the multiple source datasets might have completely different distributions and data generating processes from each other, and from the (multiple) test datasets. As mentioned, typically in transfer learning the problems have the same dimensionality or at least the same type of data (text, images, etc.), whereas our experiments consist of varied data types. Unlike many transfer learning methods, we do not use the features from test data during training.

Our work also relates to supervised learning questions of meta-learning, sometimes referred to as auto-ml (e.g., [16, 5]), learning to learn (e.g., [18]), Bayesian optimization (e.g., [15]) and lifelong learning (e.g., [17, 3]). In the case of supervised learning, where accuracy is easy to evaluate, meta-learning enables algorithms to achieve accuracy more quickly with less data. We argue that for unsupervised learning, the meta approach offers a principled means of defining and evaluating unsupervised learning.

## 3 Learning Preliminaries

A learning task consists of a universe $\mathcal{X}$, labels $\mathcal{Y}$, outputs $\mathcal{Z}$, and a bounded *loss function* $\ell : \mathcal{X} \times \mathcal{Y} \times \mathcal{Z} \to [0,1]$. A learner $L : (\mathcal{X} \times \mathcal{Y})^* \to \mathcal{Z}^{\mathcal{X}}$ takes a training set $T = (X_1, Y_1), \ldots (X_n, Y_n)$ consisting of a finite number of iid samples from $\mu$ and outputs a classifier $L(T) \in \mathcal{Z}^{\mathcal{X}}$, which denotes the set of functions from $\mathcal{X}$ to $\mathcal{Z}$. The loss of a classifier $c \in \mathcal{Z}^{\mathcal{X}}$ is $\ell_\mu(c) = \mathrm{E}_{(X,Y)\sim\mu}[\ell(X, Y, c(X))]$, and the expected loss of $L$ is $\ell_\mu(L) = \mathrm{E}_{T\sim\mu^n}[\ell_\mu(L(T))]$.

Meta-unsupervised-learning is a special case of supervised learning. There are two subtle differences. The first is conceptual: the output of a meta-unsupervised-learning algorithm is an unsupervised algorithm (e.g., a clustering algorithm) and the training examples are entire data sets. Also note that, while we require the training data to be fully labeled (e.g., clustered/classified), we may never see the true clusters of any problem encountered after deployment. This is different than online learning, where it is assumed that for each example, after you make a prediction, you find out the ground truth. The second difference is that $\mathcal{Z} \neq \mathcal{Y}$, the output of the algorithm may or may not be of the same "type" as the ground truth labels. For instance, in feature selection the output may be a subset of the features whereas the training data may have classification labels, and the loss function connects the two.

Learning is with respect to a *concept class* $\mathcal{C} \subseteq \mathcal{Z}^{\mathcal{X}}$.

**Definition 1** (Agnostic learning of $\mathcal{C}$). *For countable sets[3] $\mathcal{X}, \mathcal{Y}, \mathcal{Z}$ and $\ell : \mathcal{X} \times \mathcal{Y} \times \mathcal{Z} \to [0,1]$, a learner $L$ agnostically learns $\mathcal{A} \subseteq \mathcal{Z}^{\mathcal{X}}$ if there exists a polynomial $p$ such that for any distribution $\mu$ over $\mathcal{X} \times \mathcal{Y}$ and for any $n \geq p(1/\epsilon, 1/\delta)$,*

$$\Pr_{T\sim\mu^n} \left[ \ell_\mu(L(T)) \leq \min_{c \in \mathcal{C}} \ell_\mu(c) + \epsilon \right] \geq 1 - \delta.$$

---

[3]For simplicity of presentation, we assume that these sets are countable, but with appropriate measure theoretic assumptions the analysis in this paper extends to infinite cases in a straightforward manner.

*Further, L and the classifier L(T) must run in time polynomial in the length of their inputs.*

PAC learning refers to the special case where an additional assumption is placed on $\mu$ such that $\min_{c \in \mathcal{C}} \ell_\mu(c) = 0$. A learner $L$ is called *proper* if it only outputs classifiers in $\mathcal{C}$. We call the task *homogeneous* if $\mathcal{Y} = \mathcal{Z}$. (Most tasks we consider will be homogeneous.)

### 3.1 Meta-unsupervised-learning definitions

Meta-unsupervised-learning, which is henceforth the focus of this paper, simply refers to case where $\mu$ is a *meta-distribution* over datasets $X \in \mathcal{X}$ and ground truth labelings $Y \in \mathcal{Y}$, and classifiers $c$ are unsupervised learning algorithms that take an entire dataset $X$ as input, such as clustering algorithms. Unlike online learning, as mentioned, true labels are only observed for the training datasets.

#### 3.1.1 Meta-clustering

For a finite set $S$, the clusterings $\Pi(S)$ denotes the set of disjoint partitions of $S$ into 2 or more sets, e.g., $\Pi(\{1,2,3\}) = \{\{\{1\},\{2,3\}\}, \{\{2\},\{1,3\}\}, \{\{1,2\},\{3\}\}\}$. For a clustering $C$, denote by $\cup C = \cup_{S \in C} S$ the set of points clustered. We assume that each $X \in \mathcal{X}$ is a finite set of two or more elements and

$$\mathcal{Y} = \mathcal{Z} = \{\Pi(X) \mid X \in \mathcal{X}\}.$$

The loss function defined below is 1 if the clusterings are invalid.[4] To define the loss, for clustering $C$, we first define the distance function $d_C(x, x')$ to be 0 if they are in the same cluster, i.e., $x, x' \in S$ for some $S \in C$, i.e., and 1 otherwise. The loss will be the following clustering distance,

$$\ell(X, Y, Z) = \begin{cases} \dfrac{1}{|X|(|X|-1)} \displaystyle\sum_{x,x' \in X} |d_Y(x,x') - d_Z(x,x')| & \text{if } Y, Z \in \Pi(X) \\ 1 & \text{otherwise} \end{cases} \quad (1)$$

where $|S|$ denotes the size of set $S$. In words, this is the fraction of pairs of distinct points where the two clusterings differ on whether or not the two points are in the same cluster. Note that this loss is $1 - \text{RI}(Y, Z)$, where RI is the so-called *Rand Index*, one of the most common measures of clustering accuracy with respect to a ground truth. In our experiments, the loss we will measure is the standard *Adjusted Rand Index* (ARI) which attempts to correct the Rand Index by accounting for chance agreement [7]. We denote by $ARI(Y, Z)$ the adjusted rand index between two clusterings $Y$ and $Z$. We abuse notation and also write $ARI(Y, Z)$ when $Y$ is a vector of class labels, by converting it to a per-class clustering with one cluster for each class label.

We refer to the problem of clustering into $k=2$ clusters as *2-clustering*.

In *Euclidean clustering*, the points are Euclidean, so each dataset $X \subset \mathbb{R}^d$ for some $d \geq 1$. Note that different datasets may have different dimensionalities

---

[4]It would be natural to impose the requirement that $\cup Y = X$ with probability 1 over $\mu$, but this is not formally necessary as all clustering algorithms $c \in \mathcal{C}$ have $\cup c(X) = X$ and hence incur loss of 1 in the case that $\cup Y \neq X$.

*d. Other frameworks could be modeled*, e.g., in *list clustering* (see, e.g., [2]) where the algorithm outputs a bounded list of clusterings $Z = (Z_1, \ldots, Z_l)$, the loss could be $\min_i d(Y, Z_i)$.

Rand Index measures clustering quality with respect to an *extrinsic* ground truth. In many cases, such a ground truth is unavailable, and an *intrinsic* metric is useful. Such is the case when choosing the number of clusters. Given different clusterings of size $k = 2, 3, \ldots$, how can one compare and select? One approach is the so-called *silhouette score* [12], defined as follows for a Euclidean clustering:

$$\mathrm{sil}(C) = \frac{1}{|\cup C|} \sum_{x \in \cup C} \frac{b(x) - a(x)}{\max\{a(x), b(x)\}}, \qquad (2)$$

where $a(x)$ denotes the average distance between point $x$ and other points in its own cluster and $b(x)$ denotes the average distance between $x$ and points in an alternative cluster, where the alternative cluster is the one whose minimum average distance to $x$ is smallest among those clusters different than the one containing $x$.

## 4 Meta-unsupervised-ERM

The simplest approach to meta-unsupervised-learning is Empirical Risk Minimization (ERM), namely choosing the unsupervised algorithm from some family $\mathcal{C}$ with lowest empirical error on training set $T$, which we write as $\mathrm{ERM}_\mathcal{C}(T)$. Standard learning bounds imply a logarithmic dependence on the size of the family.

**Lemma 1.** *For any finite family $\mathcal{C}$ of unsupervised learning algorithms, any distribution $\mu$ over problems $X, Y \in \mathcal{X} \times \mathcal{Y}$, and any $m \geq 1, \delta > 0$,*

$$\Pr_{T \sim \mu^n} \left[ \ell_\mu(\mathrm{ERM}_\mathcal{C}(T)) \leq \min_{c \in \mathcal{C}} \ell_\mu(c) + \sqrt{\frac{2}{n} \log \frac{|\mathcal{C}|}{\delta}} \right] \geq 1 - \delta,$$

*where $\mathrm{ERM}_\mathcal{C}(T) \in \arg\min_{c \in \mathcal{C}} \sum_T \ell(Y, c(X))$ is any empirical loss minimizer over $c \in \mathcal{C}$.*

This standard generalization bound, which follows from Chernoff and union bounds, suffices to bound the performance of ERM over finite sets of unsupervised learning algorithms.

**Selecting among algorithms.** An immediate corollary is that, given $m$ clustering algorithms, one can perform within $O(\sqrt{\log(m)/n})$ as the best by evaluating all clustering algorithms on a training set of $n$ clustering problems (with ground truths) and choosing the best.

**Fitting a parameter.** Next, consider choosing the threshold parameter of a single linkage clustering algorithm. Fix the set of possible vertices $\mathcal{V}$. Take the universe $\mathcal{X}$ to consist of undirected weighted graphs $X = (V, E, W)$ with vertices $V \subseteq \mathcal{V}$, edges $E \subseteq \{\{u, v\} \mid u, v \in V\}$ and non-negative weights $W : E \to \mathbb{R}_+$. Data in $\mathbb{R}^d$ can be viewed as a complete graph in with $W(\{x, x'\}) = \|x - x'\|$.

The prediction and output sets are simply $\mathcal{Y} = \mathcal{Z} = \Pi(\mathcal{V})$ the sets of partitions of vertices, and the loss is defined as in Eq. (1). The classic single-linkage clustering algorithm with parameter $r \geq 0$, $L_r(V, E, w)$, creates as clusters the connected components of the subgraph of $(V, E)$ consisting of all edges whose weights are less than or equal to $r$. That is, $u, v \in V$ are in the same cluster if and only if there is a path from $u$ to $v$ where all edges in the path have weights at most $r$.

It is trivial to see that one can find the best cutoff for $r$ in polynomial time: for each edge weight $r$ in the set of edge weights across all graphs, compute the mean loss of $L_r$ across the training set. Since $L_r$ runs in polynomial time, loss can be computed in polynomial time, and the number of different possible cutoffs is bounded by the number of edge weights which is polynomial in the input size, the entire algorithm runs in polynomial time.

For a quasilinear-time algorithm (in the input size $|T| (= \Theta(\sum_i |V_i|^2))$, run Kruskal's algorithm on the union graph of all of the graphs in the training set (i.e., the number of nodes and edges are the sum of the number of nodes and edges in the training graphs, respectively). As Kruskal's algorithm adds each new edge to its forest (in order of non-decreasing edge weight), effectively two clusters in some training graph $(V_i, E_i, W_i)$ have been merged. The change in loss of the resulting clustering can be computed from the loss of the previous clustering in time proportional to the product of the two clusters that are being merged, since these are the only values on which $Z_i$ changed. Naively, this may seem to take order $\sum_i |V_i|^3$. However, note that, each pair of nodes begins separately and is updated, exactly once during the course of the algorithm, to be in the same cluster. Hence, the total number of updates is $O(\sum_i |V_i|^2)$, and since Kruskal's algorithm is quasilinear time itself, the entire algorithm is quasilinear. For correctness, it is easy to see that as Kruskal's algorithm runs, $L_r$ has been computed for each possible $r$ at the step just preceding when Kruskal adds the first edge whose weight is greater than $r$.

For generalization bounds, let us simply assume that numbers are represented with some constant $b$ number of bits, as is common on all modern computers. By Lemma 1, we see that with $m$ training graphs and $|\{L_r\}| \leq 2^b$, we have that with probability $\geq 1 - \delta$, the error of ERM is within $\sqrt{2(b + \log 1/\delta)/n}$ of $\min_r \ell_\mu(L_r)$. Combining this argument with the efficient algorithm, proves:

**Theorem 1.** *The class $\{L_r \mid r > 0\}$ of single-linkage algorithms with threshold $r$ (where numbers are represented using $b$ bits), can be agnostically learned. In particular, a quasilinear time algorithm achieves error $\leq \min_r \ell_\mu(L_r) + \sqrt{2(b + \log 1/\delta)/n}$, with probability $\geq 1 - \delta$ over the $n$ training problems.*

### 4.1 Meta-K: choosing the number of clusters, $k$

We now discuss a meta approach to choosing the number of clusters, a central problem in clustering. we refer to this as the Meta-K problem: selecting among the output of a *base* clustering algorithm that outputs one clustering with $k$ clusters for each $k \in \{2, 3, \ldots, K\}$ in a bounded range. Clearly, we require additional information as different clustering problems require different numbers of clusters. This additional "meta-information" includes information specific to the clustering and to the problem. In particular, suppose that each problem $X$ encodes, among other things, a set of points $V$ to be clustered and problem metadata $\phi \in \Phi$. Also, suppose that the base algorithm produces clusterings $(C_2, \gamma_2), \ldots, (C_K, \gamma_K) \in \mathcal{C}_V \times \Gamma$ where $C_k$ has $k$ clusters and $\gamma_k$ is meta-information about the clustering. For instance, one common type of meta-information would be some clustering quality heuristic such as the silhouette score for a clustering as defined in eq. (2). However, instead of simply choosing the clustering maximizing silhouette score, we can learn how to choose the best clustering based on $k$, $\phi$, and $\gamma_k$.

Given a family $\mathcal{F}$ of functions $f : \Phi \times \Gamma^{K-1} \to \{2, \ldots, K\}$ that select the number of clusters, as long as $\mathcal{F}$ can be parametrized by a fixed number of $b$-bit numbers, the ERM approach of choosing the "best" $f$ will be statistically efficient. If, for training problem $X_i$, the metadata is $\phi_i$ and clustering algorithm outputs $(C_{i2}, \gamma_{i2}), \ldots, (C_{iK}, \gamma_{iK})$, then ERM amounts to:

$$\text{ERM}_\mathcal{F} = \arg\min_{f \in \mathcal{F}} \sum_i \ell\left(X_i, Y_i, C_{if(\phi_i, \gamma_{i2}, \ldots, \gamma_{iK})}\right).$$

The final clustering algorithm output would be $\text{ERM}_f$ run among the output of the base clustering algorithm. If $\text{ERM}_f$ cannot be computed exactly within time constraints, an approximate minimizer may be used.

### 4.2 Meta-outlier-removal

For simplicity, we consider learning the single hyperparameter of the fraction of examples, furthest from the mean, to remove. In particular, suppose training problems are classification instances, i.e., $X_i \in \mathbb{R}^{d_i \times m_i}$ and $Y_i \in \{1, 2, \ldots, k_i\}^{m_i}$. This problem could apply to clustering or any other unsupervised learning algorithm. For this section, we assume that the base classifier algorithm (e.g., clustering algorithm) takes a parameter $\theta$ which is how many outliers to ignore during fitting.

For instance, in clustering, given an algorithm $C$, one might define $C_\theta$ with parameter $\theta \in [0, 1)$ on data $x_1, \ldots, x_n \in \mathbb{R}^d$ as follows:

1. Compute the data mean $\mu = \frac{1}{n} \sum_i x_i$.

2. Set aside as outliers the $\theta$ examples where $x_i$ is furthest from $\mu$ in Euclidean distance.

3. Run the base clustering algorithm $C$ on the data with outliers removed.

4. Assign each outlier to the cluster whose center it is closest to.

We can then compute the loss of the resulting output $Z$ and the ground truth $Y$, for any dataset. Given a meta-dataset of $n$ datasets, we choose $\theta$ so as to optimize performance. With a single $b$-bit parameter $\theta$, Lemma 1 easily implies that this choice of $\theta$ will give a loss within $\sqrt{2(b + \log 1/\delta)/n}$ of the optimal $\theta$, with probability $\geq 1 - \delta$ of the sample of datasets. The number of $\theta$'s that need to be considered is bounded by the total number of inputs across problems, so the meta-algorithm runs in polynomial time.

### 4.3 Problem recycling

For this model, suppose that each problem belongs to a set of common problem categories, e.g., digit recognition, sentiment analysis, image classification among the thousands of classes of ImageNet [13], etc. The idea is that one can recycle the solution to one version of the problem in a later incarnation. For instance, suppose that one trained a digit recognizer on a previous problem. For a new problem, the input may be encoded differently (e.g., different image size, different pixel ordering, different color representation), but there is a transformation $T$ that maps this problem into the same latent space as the previous problem so that the prior solution can be re-used.

In particular, for each problem category $i = 1, 2, \ldots, N$, there is a latent problem space $\Lambda_i$ and a solver $S_i : \Lambda_i \to \mathcal{Z}_i$. Each problem $X, Y$ of this category can be transformed to $T(X) \in \Lambda_i$ with low solution loss $\ell(X, Y, S(T(X)))$. Note that one of these categories could be "miscellaneous," a catch-all category whose solution could be another unsupervised (or meta-unsupervised) algorithm.

A problem recycling model consists of three components:

- Solvers $S_i : \Lambda_i \to \mathcal{Z}_i$, for $i = 1, 2, \ldots, N$, where each solver operates over a latent space $\Lambda_i$.

- A (meta)classifier $M : \mathcal{X} \to \{1, 2, \ldots, N\}$ that, for a problem $X$, identifies which solver $i = M(X)$ to use.

- Transformation procedures $T_i : M^{-1}(i) \to L_i$ that maps any $X$ such that $M(X) = i$ into latent space $\Lambda_i$.

The output of the meta-classifier is simply $S_{M(X)}(T_{M(X)}(X))$. Lemma 1 implies that if one can optimize over meta-classifiers and the parameters of the meta-classifier are represented by $D$ $b$-bit numbers, then one achieves loss within $\epsilon$ of the best meta-classifier with $m = O\left(Db/\epsilon^2\right)$ problems.

## 5 The *possibility* of meta-clustering

In this section, we point out how the framing of meta-clustering circumvents Kleinberg's impossibility theorem for clustering [9]. To review, Kleinberg considers clustering finite sets of points $X$ endowed with symmetric distance functions $d \in D(X)$, where the set of valid distance functions is:

$$D(X) = \{d : X \times X \to \mathbb{R} \mid \forall x, x' \in X \; d(x, x') = d(x', x) \geq 0, d(x, x') = 0 \text{ iff } x = x'\}.$$

A clustering algorithm $A$ takes a distance function $d \in D(X)$ and returns a partition, i.e., $A(d) \in \Pi(X)$.

He defines the following three axioms that should hold for any clustering algorithm $A$:

1. **Scale-Invariance**. For any distance function $d$ and any $\alpha > 0$, $A(d) = A(\alpha \cdot d)$, where $\alpha \cdot d$ is the distance function $d$ scaled by $\alpha$.

2. **Richness**. For any finite $X$ and clustering $C \in \Pi(X)$, there exists a distance $d \in D(X)$ such that $A(d) = C$.

3. **Consistency**. Let $d, d' \in D(X)$ such that $A(d) = C$, and for all $x, x' \in X$, if $x, x'$ are in the same cluster in $C$ then $d'(x, x') \leq d(x, x')$ while if $x, x'$ are in different clusters in $C$ then $d'(x, x') \geq d(x, x')$. Then $A(d') = A(d)$.

The key difficulty to defining clustering is to establish a scale. To get intuition for why, consider clustering two points $X = \{x_1, x_2\}$ so there is a single distance in the problem. For a moment, consider allowing the trivial clustering into a single cluster, so there are two legal clusterings $\{\{x_1, x_2\}\}$ and $\{\{x_1\}, \{x_2\}\}$. To see that axioms 1 and 2 are impossible to satisfy note that, by richness, there must be some distances $\delta_1, \delta_2 > 0$ such that if $d(x_1, x_2) = \delta_1$ then they are in the same cluster while if $d(x_1, x_2) = \delta_2$ they are in different clusters. This, however violates Scale-Invariance, since the two problems are at a scale $\alpha = \delta_2/\delta_1$ of each other. Now, this example fails to hold when there are two or more clusters, but it captures the essential intuition.

Now, suppose we define the clustering problem with respect to a non-empty training set of clustering problems. So a meta-clustering algorithm takes $t \geq 1$ training clustering problems $M(d_1, C_1, \ldots, d_t, C_t) = A$ with their ground-truth clusterings (on corresponding sets $X_i$, i.e., $d_i \in D(X_i)$ and $C_i \in \Pi(X_i)$) and outputs a clustering algorithm $A$. We can use these training clusterings to establish a scale.

In particular, we will show a meta-clustering algorithm whose output $A$ always satisfies the second two axioms and which satisfies the following variant of Scale-Invariance:

1. **Meta-Scale-Invariance**. Fix any distance functions $d_1, d_2, \ldots, d_t$ and ground truth clusterings $C_1, \ldots, C_t$ on sets $X_1, \ldots, X_t$. For any $\alpha > 0$, and any distance function $d$, if $M(d_1, C_1, \ldots, d_t, C_t) = A$ and $M(\alpha \cdot d_1, C_1, \ldots, \alpha \cdot d_t, C_t) = A'$, then $A(d) = A'(\alpha \cdot d)$.

With meta-clustering, the scale can be established using training data.

**Theorem 2.** *There is a meta-clustering algorithm that satisfies Meta-Scale-Invariance and whose output always satisfies Richness and Consistency.*

*Proof.* There are a number of such clustering algorithms, but for simplicity we create one based on single-linkage clustering. Single-linkage clustering satisfies Richness and Consistency (see [9], Theorem 2.2). The question is how to choose it's single-linkage parameter. One can choose it to be the minimum distance between any two points in different clusters across all training problems. It is easy to see that if one scales the training problems and $d$ by the same factor $\alpha$, the clusterings remain unchanged, and hence the meta-clustering algorithm satisfies meta-scale-invariance. □

## 6 Experiments

We conducted several experiments to substantiate the efficacy of the proposed framework under various unsupervised settings. We downloaded all classification datasets from OpenML[5] that had at most 10,000 instances, 500 features, 10 classes, and no missing data to obtain a corpus of 339 datasets. For our purposes, we extracted the numeric features from all these datasets ignoring the categorical features. We now describe in detail the results of our experiments, highlighting the gains achieved due to our meta paradigm.

---

[5] http://www.openml.org .

## 6.1 Selecting an algorithm

As a first step, we consider the question of which of a number of given clustering algorithms to use to cluster a given data set. In this section, we focus on $k = 2$ clusters. Later, we consider meta-algorithms for choosing the number of clusters.

First, one can run each of the algorithms on the repository and see which algorithm has the lowest average error. Error is calculated with respect to the ground truth labels by the ARI (see Section 3). We compare algorithms on the 250 openml binary classification datasets with at most 2000 instances. The ten base clustering algorithms were chosen to be five clustering algorithms from scikit-learn (K-Means, Spectral, Agglomerative Single Linkage, Complete Linkage, and Ward) together with a second version of each in which each attribute is first normalized to have zero mean and unit variance. Each algorithm is run with the default scikit-learn parameters.

Beyond simply choosing the algorithm that performs best across the problems, we can learn to choose a different algorithm for each problem. In order to do this, we begin with problem meta-features such as the number of features and examples in the datasets, the maximum and minimum singular values of the covariance matrix of the unlabeled data. In fact, we can use even more information. In particular, the meta-clustering algorithm for this section runs each of ten clustering algorithms on the dataset and computes what is known as a Silhouette score (see Section 3), which is a standard heuristic for evaluating clustering performance. However, rather than simply choosing the clustering with the best Silhouette score, the meta-clustering algorithm learns to select a clustering based on these scores.

To formally define the algorithm, given a clustering $\Pi$ of a dataset $X \in \mathbb{R}^{d \times m}$, let the feature vector $\Phi(X, \Pi)$ consist of the dimensionality, number of examples, minimum and maximum eigenvalues of the covariance matrix, and the silhouette score of the clustering $\Pi$:

$$\Phi(X, \Pi) = (d, m, \sigma_{\min}(\Sigma(X)), \sigma_{\max}(\Sigma(X)), \text{sil}(\Pi)),$$

where $\Sigma(X)$ denotes the covariance matrix of $X$, and $\sigma_{\min}(M)$ and $\sigma_{\max}(M)$ denote the minimum and maximum eigenvalues, respectively, of matrix $M$.

Let the training datasets be $\left\langle X_i \in \mathbb{R}^{d_i \times m_i}, Y_i \in \{0,1\}^{m_i} \right\rangle_{i=1}^{n}$. Let the base clustering algorithms be $C_j$,. The first phase of the algorithm learns a weight vector $w_j \in \mathbb{R}^5$ for each algorithm $j$. To find $w_j$, one can solve, for instance, a simple least squares linear regression[6],

$$w_j = \arg\min_{w \in \mathbb{R}^5} \sum_{i=1}^{n} \left( w \cdot \Phi(X_i, C_j(X_i)) - ARI(Y_i, C_j(X_i)) \right)^2.$$

To cluster a new dataset $X \in \mathbb{R}^{d \times m}$, the meta-algorithm then computes ARI estimates, $a_j = w_j \cdot \phi(X, C_j(X))$. It then takes the $j$ with maximal $a_j$ and outputs the clustering $C_j(x)$.

As a small modification, for each of our base algorithms, we run them a second time normalizing the data $X$ first to have mean 0 and variance 1 in each coordinate, again generating ARI estimates and taking the one with greatest estimated ARI.

---

[6]We instead used the standard $\nu$-SVR formulation for our experiments.

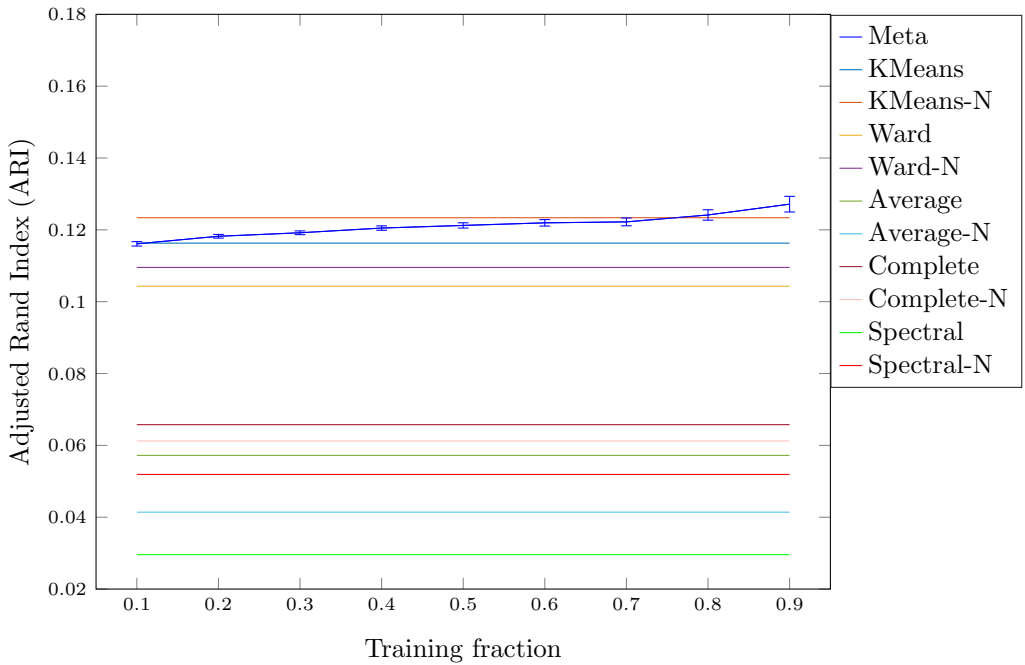

Figure 2: Adjusted Rand Index (ARI) scores of different clustering ($k$=2) algorithms on 250 openml binary classification problems. The meta algorithm is compared with standard clustering baselines on both the original data as well as the transformed data where all features in the datasets were normalized to have zero mean and unit variance (denoted by the suffix "-N"). The figure shows 95% confidence intervals for the meta approach.

The results, shown in Figure 2, demonstrate two points of interest. First, one can see that the different baseline clustering algorithms had very different average performances. Hence, a practitioner with unlabeled data (and hence little means to evaluate the different algorithms), may very likely choose a random algorithm and suffer poor performance. If one had to choose a single algorithm, K-means with normalization performed best. Perhaps this is not surprising as K-means is arguably the most popular clustering algorithm, and normalization is a common preprocessing step. However, our experiment justifies this choice based on evaluation across problems from the openml repository.

Second, Figure 2 also demonstrates that the meta-algorithm, given sufficiently many training problems, is able to outperform, on average, all the baseline algorithms. This is despite the fact that the 250 problems have different dimensionalities and come from different domains.

### 6.2 Meta-K

For the purposes of this section, we fix the clustering algorithm to be K-means and compare two approaches to choosing the number of clusters, $k$ from $k = 2$ to 10. First, we consider a standard heuristic for the baseline choice of $k$: for each cluster size $k$ and each dataset, we generate 10 clusterings from different random starts for K-means and take one with best Silhouette score among the 10. Then, over the 9 different values of $k$, we choose the one with greatest Silhouette score so that the resulting clustering is the one of greatest Silhouette score among all 90.

Similar to the approach for choosing the best algorithm above, in our meta-K approach, we learn to choose $k$ as a function of Silhouette score and $k$ by choosing the $k$ with highest estimated ARI. As above, for any given problem and any given $k$, among the 10 clusterings, we choose the one that maximizes Silhouette score. However, rather than simply choosing $k$ that maximizes Silhouette score, we choose $k$ to maximize a linear estimate of ARI that varies based on both the Silhouette score and $k$. We evaluate on exactly the same 90 clusterings for of the 339 datasets as the baseline.

We computed the Silhouette scores and Adjusted Rand Index (ARI) scores for each of the 339 datasets from $k = 2$ to 10. We conducted 10 such independent experiments for each dataset to account for statistical significance, and thus obtained two 90-dimensional vectors per dataset for Silhouette and ARI scores. Moreover, as is standard in the clustering literature, we assumed the best-fit $k_i^*$ for dataset $i$ to be the one that yielded maximum ARI score across the different runs, which is often different from $k_i$, the number of clusters in the ground truth, i.e., the number of class labels.

The training and test sets were obtained using the following procedure. Each of the 339 datasets were designated to be either a training set or test set. The number of training sets was varied over a wide range to take values in the set $\{140, 160, \ldots, 280\}$. For each such split size, the training examples from the different sets together formed the *meta-training set*, and the remaining sets formed the *meta-test set*. Thus, 8 such (training, test) partitions were obtained corresponding to these sizes.

For any particular partition into training and test sets, we followed a regression procedure to estimate the ARI score as a function of the Silhouette score. Specifically, for each $k \in \{2, \ldots, 9\}$, we fit a separate linear regression model for ARI (target or dependent variable) using Silhouette (observation) using all the data from the meta-training set in the partition pertaining to $k$: each dataset in the meta-training set provided 10 target values, corresponding to different runs where number of clusters was fixed to $k$. The models were fit using simple least-squares linear regression. Thus, we fitted 9 single feature single output linear regression models, each of size 10*number of training sets in the meta-training set.

We then used the parameters of the regression models to predict the best $k$ separately for each dataset in the meta-test set. Specifically, for each dataset in the meta-test set, we predicted the ARI score for each $k$ and each run using the parameters of regression model for $k$. Then, for each $k$, we took its predicted score to be the max score over the different runs. We took an argmax over the max scores for different $k$ to predict $\hat{k}_i$ for the dataset $i$, i.e., the optimal number of clusters (if more than one $k$ resulted in the max score, we took the smallest $k$ as $\hat{k}_i$ to break the ties). We then computed the average root mean square error (RMSE) between $\hat{k}_i$ and $k_i^*$ over the datasets in the meta-test to quantify the discrepancy between the predicted and best values of the number of clusters.

We considered our baseline $\tilde{k}$, for each set, to be the $k$ that resulted in maximum Silhouette score. Note that this is a standard heuristic to predicting the "right" number of clusters based solely on the Silhouette score. We compare the RMSE between $\tilde{k}$ and $k^*$ to the average RMSE resulting from the meta-K procedure. We report the average RMSE and the corresponding standard deviation error-bars for both the baseline and the meta-K procedures (the latter averaged over 10 independent partitions for each $t \in T$).

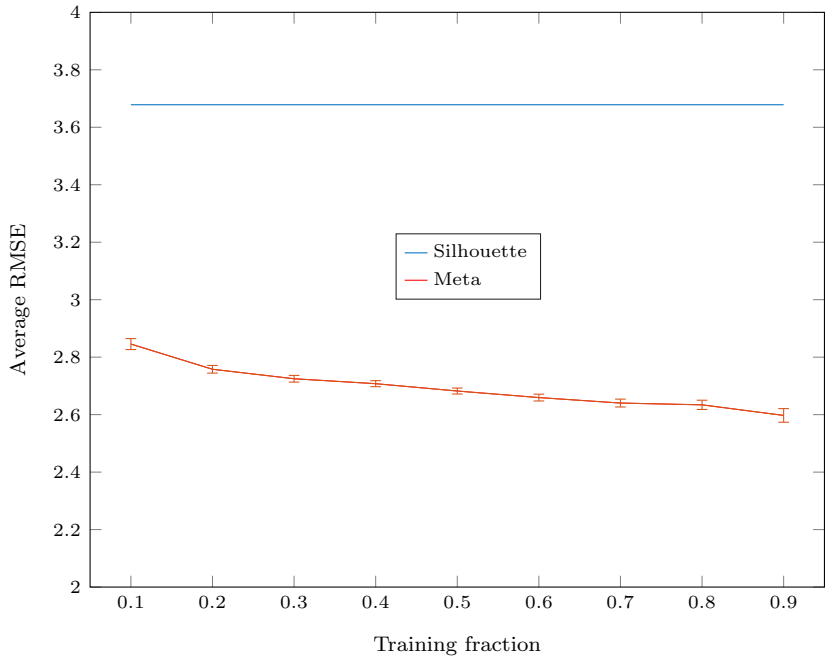

Figure 3: RMSE of distance to the best $k$, $k^*$, across datasets, i.e., comparing $\sqrt{\frac{1}{n}\sum(\tilde{k}_i - k_i^*)^2}$ and $\sqrt{\frac{1}{n}\sum(\hat{k}_i - k_i^*)^2}$. Clearly, the meta-K method outputs a number of clusters much closer to $k^*$ than the Silhouette score maximizer.

Experiments suggest that the meta-K outperforms the Silhouette baseline both in terms of the distance between the chosen $k$ and $k^*$ (Fig. 3) and in terms of ultimate loss, i.e., ARI (Fig. 4). Note again that the simple methods employed here could clearly be improved using more advance methods, but suffice to demonstrate the advantage of the meta approach.

### 6.3 Removing outliers

We also conducted experiments to validate the applicability of the meta framework to removing outliers on the same 339 binary and multi-class classification problems. Our objective was to choose a single best fraction to remove from all the meta-test sets. We now describe our experimental design for this setting. For each meta-training set $X$ in any instantiation of random partitioning obtained via the aforesaid procedure, we removed a $p \in \{0, 0.01, 0.02, \ldots, 0.05\}$ fraction examples with the highest euclidean norm in $X$ as outliers, and likewise for each meta-test set in the partition. Note that no examples were removed when $p = 0$, which is exactly the setting we considered in section 6.2. We first clustered the data without outliers, and obtained the corresponding Silhouette scores. We then put back the outliers by assigning them to their nearest cluster center, and computed the ARI score thereof. Then, following an identical procedure to the meta-K algorithm of section 6.2, we fitted regression models for ARI corresponding to complete data using the silhouette scores on pruned data, and measured the effect of outlier removal in terms of the true average ARI (corresponding to the best predicted ARI) over entire data. Again, we report the results averaged over 10 independent partitions.

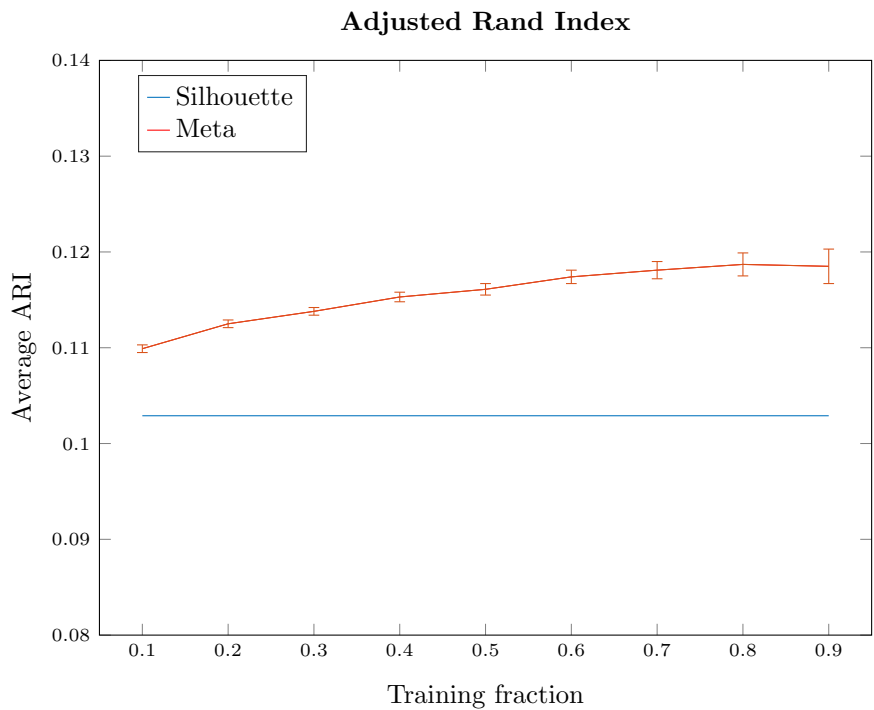

Figure 4: Average ARI scores of the meta-algorithm and the baseline for choosing the number of clusters, versus the fraction of problems used for training. We observe that the best $k$ predicted by the meta approach registered a higher ARI than that by Silhouette score maximizer. 95% Confidence interval obtained with 1000 random splits.

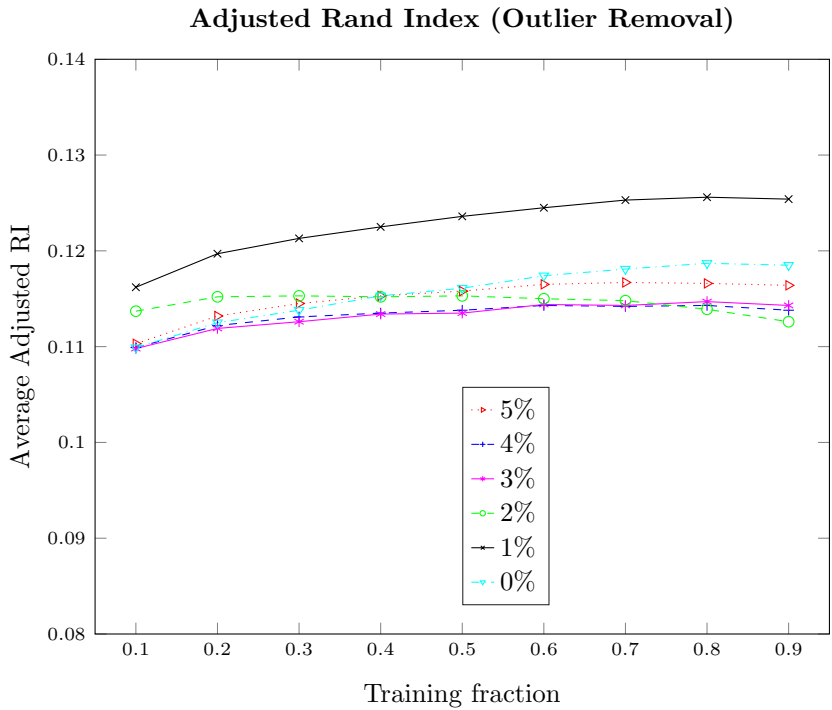

Figure 5: Outlier removal results. Removing 1% of the instances as outliers improves on the ARI scores obtained without outlier removal.

As shown in Fig. 5, by treating 1% of the instances in each dataset as outliers, we achieved an improvement in ARI scores relative to clustering with all the data as in section 6.2. As the fraction of data considered outlier was increased to 2% or higher, however, we observed a substantial dip in the performance. While 1% was the optimal fraction among the candidates, clearly further parameter search and improved outlier removal algorithms could provide additional benefit. Our goal in this paper is mainly to pose the questions and demonstrate that meta-learning can improve performance across problems from different domains and of possibly different dimensionality.

## 6.4 Deep learning a binary similarity function

In this section, we consider a new unsupervised problem of learning a binary similarity function (BSF) that predicts whether two examples from a given problem should belong to the same cluster (i.e., have the same class label). Formally, a problem is specified by a set $X$ of data and meta-features $\phi$. The goal is to learn a classifier $f(x, x', \phi) \in \{0, 1\}$ that takes two examples $x, x' \in X$ and the corresponding problem meta-features $\phi$, and predicts 1 if the input pair would belong to the same cluster (or have the same class labels). Here, we approximate this function using a neural network.

In our experiments, we take Euclidean data $X \subseteq \mathbb{R}^d$ (each problem may have different dimensionality $d$) and the meta-features $\phi = \Sigma(X)$ consist of the covariance matrix of the unlabeled data. We restricted our experiments to the 146 datasets with at most 1000 examples and 10 features. We normalized each dataset to have zero mean and unit variance along every coordinate. Hence the covariance matrix was 1 on the diagonal.

We randomly sampled pairs of examples from each dataset to form *meta-training* and *meta-test* sets as described in the following section 6.4.1. For each pair, we concatenated the features to create data with 20 features (padding examples with fewer than 10 features with zeros). We then computed the empirical covariance matrix of the dataset, and vectorized the entries of the covariance matrix on and above the leading diagonal to obtain an additional 55 covariance features. We then concatenated these features with the 20 features to form a 75-dimensional feature vector per pair. Thus all pairs sampled from the same dataset shared the 55 covariance features. Moreover, we derived a new binary label dataset in the following way. We assigned a label 1 to pairs formed by combining examples belonging to the same class, and 0 otherwise. In summary, each of the 146 datasets resulted in a new dataset with 75 features and binary labels.

### 6.4.1 Sampling pairs to form meta-train and meta-test datasets

We formed a partition of the new datasets by randomly assigning each dataset to one of the two categories with equal probability. Each dataset in the first category was used to sample data pairs for both the meta-training and the meta-internal_test (meta-IT) datasets, while the second category did not contribute any training data and was exclusively used to generate only the meta-external_test (meta-ET) dataset.

We constructed meta-training pairs by sampling randomly pairs from each dataset in the first category. In order to mitigate the bias resulting from the variability in size of the different datasets, we restricted the number of pairs sampled from each dataset to at most 2500. Likewise, we obtained the meta-IT dataset by collecting randomly sampling each dataset subject to the maximum 2500 pairs. Specifically, we randomly shuffled each dataset belonging to the first category, and used the first half (or 2500 examples, whichever was fewer) of the dataset for the meta-training data, and the following indices for the meta-IT data, again subject to maximum 2500 instances. This procedure ensured a disjoint intersection between the meta-training and the meta-IT data. Note that combining thousands of examples from each of hundreds of problems yields hundreds of thousands of examples, **turning small data into big data**. This provides a means of making DNNs naturally applicable to data sets that might have otherwise been too small.

We created the meta-ET data using datasets belonging to the second category. Again, we sampled at most 2500 examples from each dataset in the second category. We emphasize that the datasets in the second category did not contribute any training data for our experiments.

We performed 10 independent experiments to obtain multiple partitions of the datasets into two categories, and repeated the aforementioned procedure to prepare 10 separate (meta-training, meta-IT, meta-ET) triplets. This resulted in the following (average size +/- standard deviation) statistics for dataset sizes:

$$\text{meta-training and meta-IT} \;:\; 173,762 \pm 10,739$$
$$\text{meta-ET} \;:\; 172,565 \pm 11,915$$

In order to ensure symmetry of the binary similarity function, we introduced an additional meta-training pair for each meta-training pair in the meta-training set: in this new pair, we swapped the order of the feature vectors of the two instances while replicating the covariance features of the underlying dataset that contributed the two instances (the covariance features, being symmetric, carried over unchanged).

### 6.4.2 Training neural models

For each meta-training set, we trained an independent deep net model with 4 hidden layers having 100, 50, 25, and 12 neurons respectively over just 10 epochs, and used batches of size 250 each. We updated the parameters of the model via the Adadelta[7] implementation of the stochastic gradient descent (SGD) procedure supplied with the Torch library[8] with the default setting of the parameters, specifically, interpolation parameter equal to 0.9 and no weight decay. We trained the model via the standard negative log-likelihood criterion (NLL). We employed ReLU non-linearity at each hidden layer but the last one, where we invoked the log-softmax function.

We tested our trained models on meta-IT and meta-ET data. For each feature vector in meta-IT (respectively meta-ET), we computed the predicted same class probability. We added the predicted same class probability for the feature vector obtained with flipped order, as described earlier for the feature vectors in the meta-training set. We predicted the instances in the corresponding pair to be in the same cluster if the average of these two probabilities exceeded 0.5, otherwise we segregated them.

### 6.4.3 Results

We compared the meta approach to a hypothetical majority rule that had prescience about the class distribution.[9] As the name suggests, the majority rule predicted all pairs to have the majority label, i.e., on a problem-by-problem basis we determined whether 1 (same class) or 0 (different class) was more accurate and gave the baseline the advantage of this knowledge for each problem, even though it normally wouldn't be available at classification time. This information about the distribution of the labels was not accessible to our meta-algorithm.

---

[7]described in http://arxiv.org/abs/1212.5701 .
[8]available at: https://github.com/torch/torch7 .
[9]Recall that we reduced the problem of clustering to pairwise binary classification, where 1 represented same cluster relations.

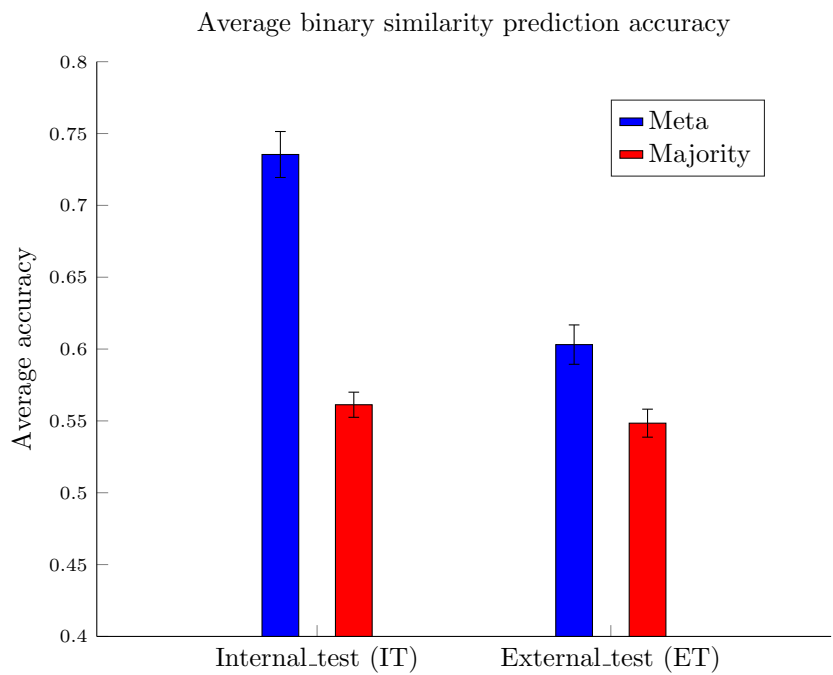

Figure 6: Mean accuracy and standard deviation on meta-IT and meta-ET data. Comparison between the fraction of pairs correctly predicted by the meta algorithm and the majority rule. Recall that meta-ET, unlike meta-IT, was generated from a partition that did not contribute any training data. Nonetheless, the meta approach significantly improved upon the majority rule even on meta-ET.

Fig. 6 shows the average fraction of similarity pairs correctly identified relative to the corresponding pairwise ground truth relations on the two test sets, and the corresponding standard deviations across the 10 independent (meta-training, meta-IT, meta-ET) collections. Clearly, the meta approach outperforms the majority rule on meta-IT, illustrating the benefits of the meta approach in a multi-task transductive setting. More interesting, still, is the significant improvement exhibited by the meta method on meta-ET, despite having its category precluded from contributing any data for training. The result clearly demonstrates the benefits of leveraging archived supervised data for informed decision making in unsupervised settings such as binary similarity prediction.

# 7 Conclusion

Currently, UL is difficult to define. We argue that this is because UL problems are generally considered in isolation. We suggest that they can be viewed as representative samples from some meta-distribution $\mu$ over UL problems. We show how a repository of multiple datasets annotated with ground truth labels can be used to improve average performance. Theoretically, this enables us to make UL problems, such as clustering, well-defined.

At a very high level, one could even consider the distribution of data run through a popular clustering toolkit, such as scikit-learn. While there is a tremendous variety of problems encountered by such a group, an individual clustering course reviews may find that they are not even the first to wish to cluster course reviews using the tool.

Prior datasets may prove useful for a variety of reasons, from simple to complex. The prior datasets may help one choose the best clustering algorithm or parameter settings, or shared features may be transferred from prior datasets that can be identified as useful. We also demonstrate how the combination of many small data sets can form a large data set appropriate for a neural network.

This work is meant only as a first step in defining the problem of meta-unsupervised-learning. The algorithms we have introduced can surely be improved in numerous ways, and the experiments are mainly intended to show that there is potential for improving performance using a heterogeneous repository of prior labeled data.

# References

[1] M. Ackerman and S. Ben-David. Measures of clustering quality: A working set of axioms for clustering. In *NIPS*, 2008.

[2] M.-F. Balcan, A. Blum, and S. Vempala. A discriminative framework for clustering via similarity functions. In *Proceedings of the Fortieth Annual ACM Symposium on Theory of Computing (STOC)*, pages 671–680, 2008.

[3] M.-F. Balcan, A. Blum, and S. Vempala. Efficient representations for lifelong learning and autoencoding. In *Workshop on Computational Learning Theory (COLT)*, 2015.

[4] H. Daumè III and D. Marcu. A bayesian model for supervised clustering with the dirichlet process prior. *Journal of Machine Learning Research (JMLR)*, 6:1551–1577, 2005.

[5] M. Feurer, A. Klein, K. Eggensperger, J. Springenberg, M. Blum, and F. Hutter. Efficient and robust automated machine learning. In *Advances in Neural Information Processing Systems*, pages 2962–2970, 2015.

[6] T. Finley and T. Joachims. Supervised clustering with support vector machines. In *ICML*, 2005.

[7] L. Hubert and P. Arabie. Comparing partitions. *Journal of classification*, 2(1):193–218, 1985.

[8] M. J. Kearns, R. E. Schapire, and L. M. Sellie. Toward efficient agnostic learning. *Machine Learning*, 17(2-3):115–141, 1994.

[9] J. Kleinberg. An impossibility theorem for clustering. *Advances in neural information processing systems*, pages 463–470, 2003.

[10] S. J. Pan and Q. Yang. A survey on transfer learning. *IEEE Transactions on Knowledge and Data Engineering (TKDE)*, 22:1345–1359, 2010.

[11] F. Pedregosa, G. Varoquaux, A. Gramfort, V. Michel, B. Thirion, O. Grisel, M. Blondel, P. Prettenhofer, R. Weiss, V. Dubourg, J. Vanderplas, A. Passos, D. Cournapeau, M. Brucher, M. Perrot, and E. Duchesnay. Scikit-learn: Machine learning in Python. *Journal of Machine Learning Research*, 12:2825–2830, 2011.

[12] P. J. Rousseeuw. Silhouettes: a graphical aid to the interpretation and validation of cluster analysis. *Journal of computational and applied mathematics*, 20:53–65, 1987.

[13] O. Russakovsky, J. Deng, H. Su, J. Krause, S. Satheesh, S. Ma, Z. Huang, A. Karpathy, A. Khosla, M. Bernstein, A. C. Berg, and L. Fei-Fei. ImageNet Large Scale Visual Recognition Challenge. *International Journal of Computer Vision (IJCV)*, 115(3):211–252, 2015.

[14] M. Sahlgren. The word-space model: Using distributional analysis to represent syntagmatic and paradigmatic relations between words in high-dimensional vector spaces. 2006.

[15] J. Snoek, H. Larochelle, and R. P. Adams. Practical bayesian optimization of machine learning algorithms. In *Advances in neural information processing systems*, pages 2951–2959, 2012.

[16] C. Thornton, F. Hutter, H. H. Hoos, and K. Leyton-Brown. Auto-weka: Combined selection and hyperparameter optimization of classification algorithms. In *Proceedings of the 19th ACM SIGKDD international conference on Knowledge discovery and data mining*, pages 847–855. ACM, 2013.

[17] S. Thrun and T. M. Mitchell. Lifelong robot learning. In *The biology and technology of intelligent autonomous agents*, pages 165–196. Springer, 1995.

[18] S. Thrun and L. Pratt. *Learning to learn*. Springer Science & Business Media, 2012.

[19] L. G. Valiant. A theory of the learnable. *Communications of the ACM*, 27(11):1134–1142, 1984.

[20] R. B. Zadeh and S. Ben-David. A uniqueness theorem for clustering. In *Proceedings of the twenty-fifth conference on uncertainty in artificial intelligence*, pages 639–646. AUAI Press, 2009.